\documentclass[10pt,twocolumn,letterpaper]{article}

\usepackage[pagenumbers]{cvpr} 

\usepackage{amsmath,amsfonts,bm}

\def\eqref#1{equation~\ref{#1}}

\def\1{\bm{1}}

\DeclareMathAlphabet{\mathsfit}{\encodingdefault}{\sfdefault}{m}{sl}
\SetMathAlphabet{\mathsfit}{bold}{\encodingdefault}{\sfdefault}{bx}{n}

\newcommand{\R}{\mathbb{R}}

\definecolor{customgreen}{RGB}{152, 206, 159}

\newcommand{\sctwo}{{AP}${}^{}_{\text{role2}}$}
\newcommand{\apf}{{Full}}
\newcommand{\apnr}{{Non-rare}}
\newcommand{\apr}{{Rare}}

\definecolor{cvprblue}{rgb}{0.21,0.49,0.74}
\usepackage[pagebackref,breaklinks,colorlinks,allcolors=cvprblue]{hyperref}

\usepackage[accsupp]{axessibility} 

\usepackage{url}
\usepackage{amsmath}
\usepackage{amssymb}
\usepackage{algorithm,algpseudocode}
\usepackage{graphicx}
\usepackage{tabularx}
\usepackage{adjustbox}
\usepackage{svg}
\usepackage{wrapfig}
\usepackage{stmaryrd}
\usepackage{booktabs}
\usepackage{caption}
\usepackage{multirow}
\usepackage{svg}
\usepackage{pifont}
\usepackage{fancybox}
\usepackage{makecell}
\usepackage{colortbl}
\usepackage{xcolor}
\usepackage{amsfonts}
\definecolor{mygray}{gray}{.95}
\usepackage{soul}

\definecolor{mygray}{gray}{.95}
\definecolor{vcoco}{RGB}{220, 230, 255}
\definecolor{hico}{RGB}{220, 255, 220}
\definecolor{hicodet}{RGB}{170, 230, 170}
\definecolor{swig}{RGB}{255, 230, 180} 
\definecolor{swig-det}{RGB}{255, 200, 120}

\newcolumntype{Y}{>{\centering\arraybackslash}p{1.4cm}}

\def\paperID{20366} 
\def\confName{CVPR}
\def\confYear{2026}

\title{RegFormer: Transferable Relational Grounding \\ for Efficient Weakly-Supervised Human-Object Interaction Detection}

\author{
Jihwan Park\textsuperscript{\rm 1} \hspace{0.5cm}
Chanhyeong Yang\textsuperscript{\rm 2}\thanks{Work done while at Korea University.} \hspace{0.5cm} 
Jinyoung Park\textsuperscript{\rm 1} \hspace{0.5cm}
Taehoon Song\textsuperscript{\rm 1} \hspace{0.5cm}
Hyunwoo J. Kim\textsuperscript{\rm 1}\thanks{Corresponding author.}\\
\textsuperscript{\rm 1}KAIST \hspace{0.5cm} \textsuperscript{\rm 2}LG Energy Solution \\
{\tt\small \{\href{mailto:jseven7071@kaist.ac.kr}{jseven7071}, \href{mailto:jinyoung.park@kaist.ac.kr}{jinyoung.park}, \href{mailto:taehoons@kaist.ac.kr}{taehoons}, \href{mailto:hyunwoojkim@kaist.ac.kr}{hyunwoojkim}\}@kaist.ac.kr \quad \href{mailto:chanhyeong_y@lgensol.com}{chanhyeong\_y}@lgensol.com}
}

\begin{document}
\maketitle
\begin{abstract}
Weakly-supervised Human–Object Interaction (HOI) detection is essential for scalable scene understanding, as it learns interactions from only image-level annotations.
Due to the lack of localization signals, prior works typically rely on an external object detector to generate candidate pairs and then infer their interactions through pairwise reasoning.
However, this framework often struggles to scale due to the substantial computational cost incurred by enumerating numerous instance pairs. 
In addition, it suffers from false positives arising from non-interactive combinations, which hinder accurate instance-level HOI reasoning.
To address these issues, we introduce \textbf{Re}lational \textbf{G}rounding Trans\textbf{former} (\textbf{RegFormer}), a versatile interaction recognition module for efficient and accurate HOI reasoning.
Under image-level supervision, RegFormer leverages spatially grounded 
signals as guidance for the reasoning process and promotes locality-aware interaction learning.
By learning localized interaction cues, our module distinguishes humans, objects, and their interactions,
enabling direct transfer from image-level interaction reasoning to precise and efficient instance-level reasoning without additional training.
Our extensive experiments and analyses demonstrate that RegFormer effectively learns spatial cues for instance-level interaction reasoning, operates with high efficiency, and even achieves performance comparable to fully supervised models.
Our code is available at \href{https://github.com/mlvlab/RegFormer}{https://github.com/mlvlab/RegFormer}.
\end{abstract}


\section{Introduction}
\label{sec:intro}

Human–Object Interaction (HOI) detection is a fundamental task in comprehensive scene understanding, which aims to localize humans and objects in an image and identify their interactions as triplets $\langle\text{human, interaction, object}\rangle$.
Typically, training HOI detection models relies on \emph{full} supervision that specifies object categories, bounding boxes, and their interaction labels.
However, as HOI datasets continue to scale, annotating interaction classes for all possible human–object pairs in the image becomes prohibitively expensive, making full annotation infeasible in practice.

\begin{figure}[t]
\centering
\includegraphics[width=0.85\columnwidth]{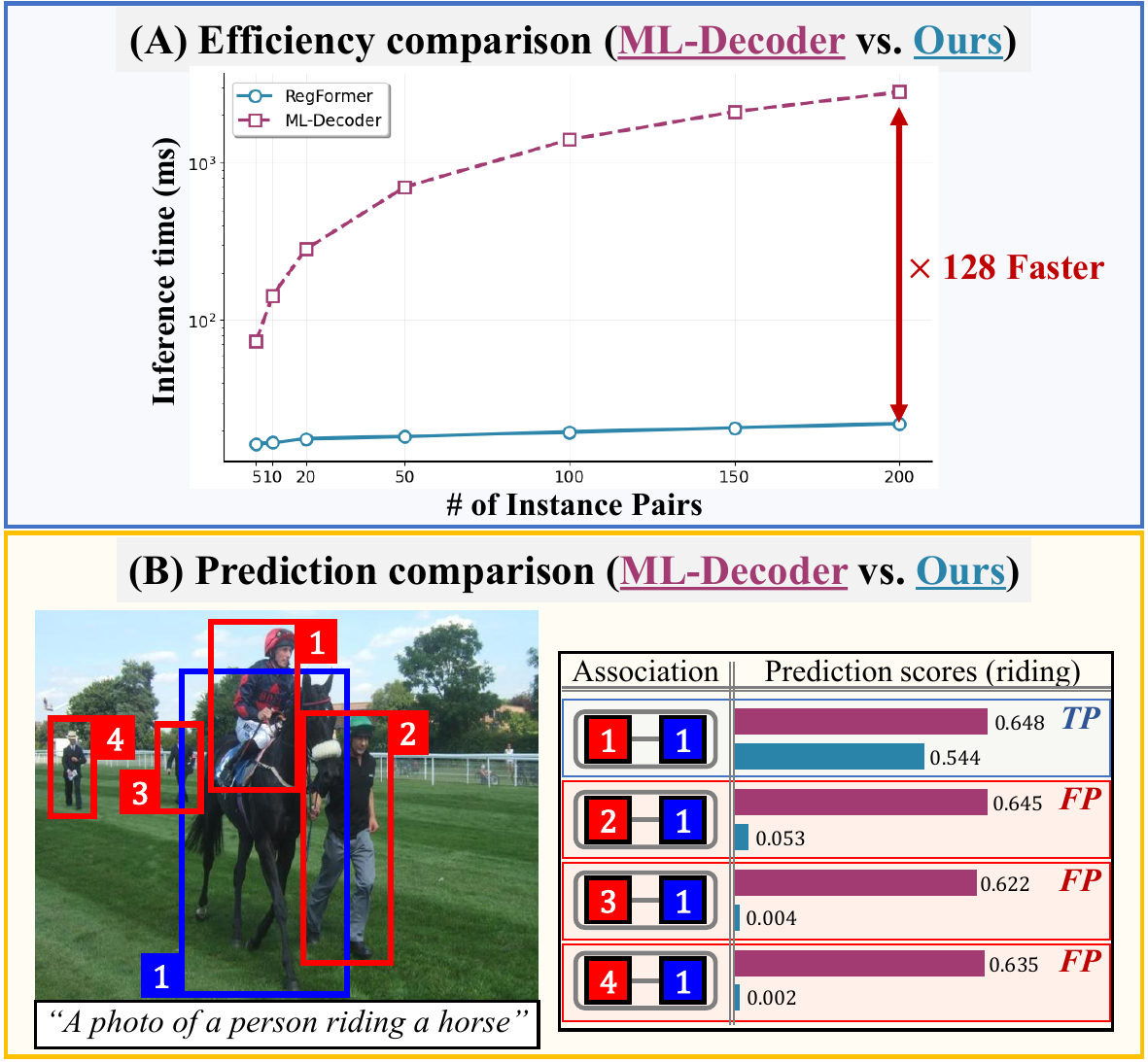}
\vspace{-0.2cm}
\caption{{\textbf{Comparison of weakly-supervised HOI detection frameworks.}} 
\textbf{(A)} As the number of instance pairs increases, RegFormer shows only a marginal increase in inference time, whereas the ML-Decoder becomes significantly slower.
\textbf{(B)} In addition, RegFormer effectively suppresses non-interactive human–object pairs, producing less false positives.}
\label{fig:comparison_fig1}
\end{figure}

To address this, recent approaches leverage {weakly supervised learning}~\cite{explainhoi,mxhoi,alignformer,vlhoi,pprfcn,weakhoiclip,opencat}, 
using only image-level annotations that indicate which HOI triplet classes are present in an image, 
without the localization of humans and objects.
In the absence of explicit local supervision, these approaches typically rely on human and object proposals generated by an off-the-shelf detector, which are paired and processed by an interaction classification module to predict pairwise interactions.
Since the detector produces numerous human–object candidates, the classification module plays a crucial role in efficiently processing them and identifying true interactions with high precision.
However, previous works~\cite{ml_decoder,RAM} often suffer from substantial computational overhead~\cite{ada_cm} (\cref{fig:comparison_fig1}-(A)), as they repeatedly crop pairwise union regions from the image and forward each cropped region through the model.
To improve efficiency, RoI-Align~\cite{maskrcnn} is often employed to extract union region features~\cite{weakhoiclip} directly from the spatial feature map of a pretrained backbone such as CLIP~\cite{CLIP}, allowing all interactions to be inferred with a single forward pass.
Nevertheless, this strategy still faces generalization issues, as union regions often include irrelevant areas that mislead classification for a specific pair (\cref{fig:comparison_fig1}-(B)).
Alternatively, instance features from off-the-shelf detectors~\cite{mxhoi,explainhoi} can directly encode the relevant humans and objects, but this design tightly couples the classifier with the detector, requiring retraining whenever the detector is replaced.

To overcome these issues, we present \textbf{Re}lational \textbf{G}rounding Trans\textbf{former} (\textbf{RegFormer}), a lightweight yet versatile interaction recognition module that unifies image-level and instance-level HOI reasoning within a single framework.
Unlike prior works that rely on exhaustive triplet queries, RegFormer performs in a \textit{sequential} manner, where it first generates queries for human–object pairs (\textbf{HO}) at \textit{pairwise instance encoder} and then predicts the corresponding interactions (\textbf{I}) in \textit{interaction decoder}.
In the pairwise instance encoder, HO queries are constructed as \textit{spatially grounded representations} by aggregating relevant features from the spatial feature maps of the human–object pairs.
This design enables the model to implicitly learn spatial cues for effective interaction reasoning.
In addition, we incorporate an implicit localization signal to learn an \textit{interactiveness score} for each human–object pair.
This score acts as an explicit ``gating'' mechanism that suppresses irrelevant objects, preventing spurious feature activation and enabling the model to focus on localized instances.

During HOI detection, human and object instances predicted by the off-the-shelf detector are used to constrain both the query construction and interactiveness scoring within the region of each specific human–object instance.
This instance-aware design enhances discriminability across different instance pairs while suppressing non-interactive combinations, 
thereby enabling direct transfer from image-level to instance-level HOI reasoning without additional training.

Extensive experiments demonstrate that RegFormer effectively models spatial relationships with high efficiency, not only outperforms prior weakly supervised approaches by a clear margin but also achieves performance competitive with fully supervised methods.
The contributions of RegFormer are summarized as:
\begin{itemize}
    \item We propose \textbf{Re}lational \textbf{G}rounding Trans\textbf{former} (\textbf{RegFormer}), a lightweight and versatile classification module that unifies image-level and instance-level HOI reasoning within a single framework.
    \item We enhance the capability of RegFormer by integrating spatially grounded query and interactiveness-aware learning, enabling the model to better capture localized interaction cues.
    \item {RegFormer} seamlessly transfers from image-level classification to instance-level detection without additional training, enabling efficient instance-level interaction reasoning that surpasses previous weakly supervised methods and competes with fully supervised counterparts.
\end{itemize}

\section{Related works}
\subsection{HOI detection}
HOI detection models are typically trained with $\langle \text{human, interaction, object} \rangle$ triplets that include object categories, interaction labels, and precise localization of the human–object instances participating in each interaction.
With such rich annotations, existing supervised methods operate either in a one-stage manner~\cite{uniondet, hotr, speaq,cpc,qpic, genvlkt, ppdm, rlip, ager, eoid, clip4hoi}, jointly predicting humans, objects, and interactions, or in a two-stage manner~\cite{upt, pvic, ada_cm, cmmp} that first detects instances and then classifies their pairwise interactions.
However, the annotation cost remains substantial because training these models requires specifying the interactions of all humans and objects appearing in each image.
Zero-shot HOI detection~\cite{fcl, atl, vcl, hoiclip, clip4hoi, cmmp,vdrp} reduces the need for exhaustive class-level supervision, but still assumes full instance-level training annotations.
These limitations motivate the study of \textit{weakly supervised HOI detection}, where human and object locations are no longer required.
\subsection{Weakly supervised HOI detection}
Weakly supervised HOI detection aims to recognize human–object interactions without exhaustive instance-level annotations, \eg, locations of all human and object instances involved in each interaction.  
Early studies~\cite{mxhoi, alignformer} explored pseudo-alignment and auxiliary learning schemes to associate humans and objects under limited supervision, while recent methods~\cite{weakhoiclip, vlhoi} leveraged vision–language models or large-scale pretraining~\cite{opencat} to enhance interaction reasoning from weak image-level signals.
However, in the absence of localization supervision, these methods depend on an off-the-shelf object detector to extract human–object instance pairs, placing the burden on the classification module to efficiently process \textit{many instance pairs} and accurately \textit{filter out non-interactive ones}.
Motivated by this, we aim to develop an effective, versatile classification module that seamlessly transfers from image-level to instance-level HOI reasoning.
Building on the successful multi-label classification design of ML-Decoder~\cite{ml_decoder,RAM}, our module is designed to satisfy both efficiency and discriminative reasoning requirements, enabling instance-level HOI inference without any additional training.

\section{Preliminaries}
In this section, we provide a brief overview of weakly supervised Human–Object Interaction~(HOI) detection and Multi-Label Decoder~(ML-Decoder), which we adopt as our base architecture due to its effectiveness in multi-label classification.

\label{sec:3.1}
\noindent\textbf{Problem setting.}
Our task is Human–Object Interaction (HOI) detection under a \textit{weakly supervised} setting, where only image-level HOI annotations~(\ie, object and action labels) are available during training, yet the model should reason over instance-level pairwise interactions at inference time.
Specifically, the model is trained on an image $I$ annotated with a set of HOI triplets $C=\{c^\text{hoi}\}$, where each triplet $c^\text{hoi}$ consists of a human, an action class $c^\text{a}$, and an object class $c^\text{o}$, \eg, \textit{human-ride-bicycle}.
At inference time, given an image $I$, the model predicts a set of localized human–object pairs and their corresponding interactions, $\tilde{P}=\{{(\tilde{b}^{{{\text{h}}}},\tilde{b}^{{\text{o}}},\tilde{c}^{{\text{o}}},\tilde{c}^\text{a},\tilde{s}^\text{a})}\}$, where $\tilde{b}^{{\text{h}}}$ and $\tilde{b}^{{\text{o}}}$ denote the predicted bounding boxes of the human and object, $\tilde{c}^{{\text{o}}}$ indicates the object class, $\tilde{c}^\text{a}$ represents the predicted action label between them, and $\tilde{s}^\text{a}$ is the action confidence score.

Due to the absence of instance-level localization in training annotations, weakly supervised HOI detectors typically follow a two-stage pipeline: an off-the-shelf object detector~\cite{detr,faster_rcnn} first detects $N_{{\tilde{\text{h}}}}$ human and $N_{\tilde{\text{o}}}$ object instances in an image, after which all possible human–object pairs are formed and passed to an interaction classification module for pairwise relation prediction.

\noindent\textbf{Base architecture.}
We build our HOI detector upon the cross-attention–based multi-label classifier architecture~(ML-Decoder)~\cite{ml_decoder,RAM}, given its remarkable effectiveness in multi-label classification.
For HOI classification, ML-Decoder uses text embeddings of HOI classes~(\eg, \textit{``human ride bicycle''}, \textit{``human hold cup''}) as queries.
Formally, for each HOI class $c^\text{hoi}_t$ and pretrained text encoder $\mathcal{T}$ (\eg, CLIP~\cite{CLIP}), the corresponding query is initialized as:
\begin{equation}
    q^\text{hoi}=\{\mathcal{T}(c^\text{hoi}_t)\texttt{W}_q\}_{t=1}^{N_\text{hoi}}\in\R^{N_\text{hoi}\times d},
\end{equation}
where $\texttt{W}_q$ is a projection matrix for the query.
These text-derived query embeddings provide rich semantic priors, allowing the model to infer unseen interactions during training.

Then, the decoder performs cross-attention with the queries and the spatial feature map $f=\mathcal{V}\left(I\right)$ extracted from the image backbone $\mathcal{V}$, to calculate a classification score $\hat{s}^\text{hoi} \in \mathbb{R}^{N_{\text{hoi}}}$:
\begin{equation}
    \bar{q}^\text{hoi} = \texttt{Att}(q^\text{hoi}, f, f) , \quad
    \hat{s}^\text{hoi} =\sigma(\bar{q}^\text{hoi}\texttt{W}_p) ,
    \label{eq:ml_decoder}
\end{equation}
where $\texttt{W}_p$ denotes projection matrix for classification score, $\sigma$ represents the sigmoid function~\cite{siglip} with temperature and bias terms, and $\texttt{Att}(\textit{query},\textit{key},\textit{value})$ is an attention module operating on query, key, and value inputs.

A straightforward approach~\cite{ada_cm} to extend ML-Decoder for HOI detection is to first obtain human and object bounding boxes, crop a union region for each detected human–object pair within the image, and then feed the resulting cropped images into the classification module.
However, this naive strategy is highly inefficient since it requires a forward pass for every $N_{{\tilde{\text{h}}}}\times N_{\tilde{\text{o}}}$  possible instance pair.
Moreover, as the scene becomes denser, the cropped union region often contains irrelevant instances, producing numerous false positives.

\section{Method}

\begin{figure*}[t!]
    \centering 
    \includegraphics[width=0.95\textwidth]{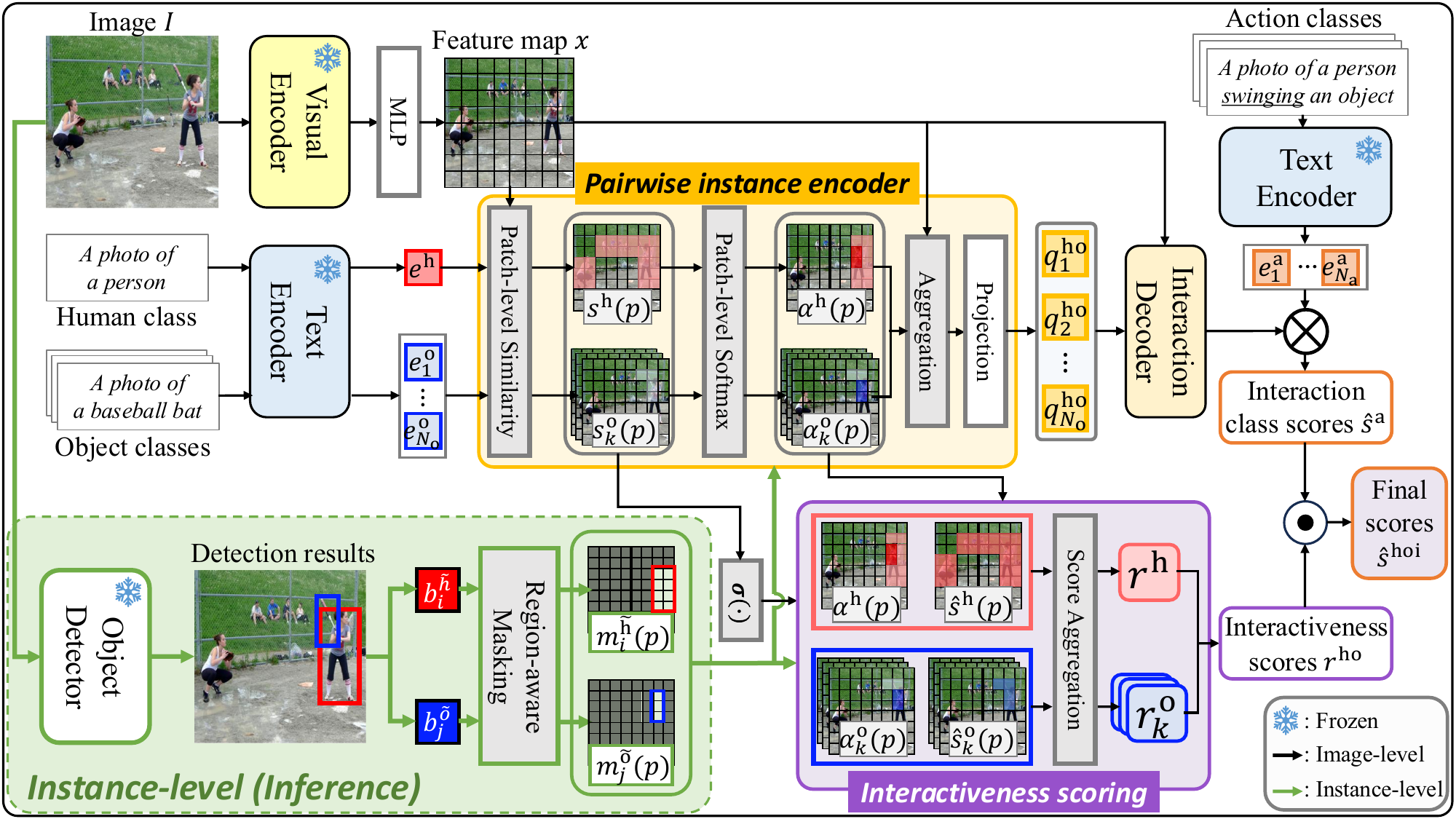}
    \vspace{-0.1cm}
    \caption{\textbf{Overall framework of RegFormer.}    
    RegFormer unifies image-level and instance-level reasoning within a single framework by learning to capture spatial cues for interaction reasoning using only image-level labels.    
    During training, \textit{pairwise instance encoder} constructs a human–object (HO) query $q^{\text{ho}}$ by aggregating spatial features $x$ based on patch importance score $\alpha(p)$ associated with each human and object class.
    The resulting HO queries are processed by \textit{interaction decoder}, which outputs the interaction classification scores $\hat{s}^{\text{a}}$.
    To further support locality-aware learning, we introduce a spatially aggregated \textit{interactiveness score} ${r}$, which acts as a gating signal for the interaction score $\hat{s}^\text{a}$ and receives explicit supervision.
    During instance-level HOI inference, given detected human and object instances from the object detector, \textit{region-aware mask} $m(p)$ is applied to constrain both the HO queries and the interactiveness scores within their corresponding regions, enabling HOI detection without additional training.
    }
    \vspace{-5pt}
\label{fig:method} 
\end{figure*}
We present \textbf{Re}lational \textbf{G}rounding Trans\textbf{former}~(RegFormer), which bridges the gap between image-level (\ie, HOI classification) and instance-level HOI reasoning (\ie, HOI detection) within a unified and efficient framework.
Since instance-level annotations, \eg, bounding boxes, are unavailable during the training phase, RegFormer generates human–object (HO) queries by injecting spatial cues of human and object with a \textit{pairwise instance encoder}, and then decodes them through the \textit{interaction decoder} to produce final interaction classification scores.
RegFormer also learns the \textit{interactiveness} of each human–object pair, providing an explicit supervisory signal that guides the model toward more effective locality-aware reasoning.
This framework enables precise yet efficient HOI detection, requiring only a simple modification to focus the module on instance-specific regions without additional training.
The overall pipeline is illustrated in \cref{fig:method}.

\subsection{RegFormer for HOI classification}
\label{sec:4.1}
Unlike prior methods~\cite{ml_decoder,RAM} that utilize a query set of all HOI classes, RegFormer adopts a \textit{sequential} framework~\cite{upt,pvic}, where queries are first grouped by human–object class pairs (\textbf{HO}) and then used to predict their corresponding interaction classes (\textbf{I}) for each pair, \ie, $\textbf{HO} \rightarrow \textbf{I}$.
{This design enables RegFormer to handle a large number of instance pairs without incurring significant computational overhead.}

\noindent\textbf{Pairwise Instance Encoder.}
The pairwise instance encoder learns human–object representations to generate localized HO queries, which are used as input for the interaction decoder.
Here, a simple way to construct HO queries is to use image-level text-driven embeddings, in which each query is initialized from textual embeddings extracted from the pre-trained text encoder $\mathcal{T}$, similar to \cite{ml_decoder}.
However, since text-based initialization lacks spatial information, incorporating local priors (\eg, box coordinates) into the queries becomes difficult without additional training~\cite{upt,pvic}.
This limits the transferability of image-level reasoning to instance-level reasoning.

To address this, we incorporate class-aware spatial cues with the patch-level similarity to generate a \textit{spatially grounded pairwise query}.
Specifically, we highlight the regions where the human and object are likely to appear and interact, and aggregate their corresponding visual features to better capture spatial information for interaction reasoning.
First, we calculate the objectiveness score on the spatial patch $p$ on the image feature map by measuring the cosine similarity between each patch feature $x(p) \in \R^d$ and the text embeddings of the human ${e}^\text{h} \in \R^{d_\text{t}}$ and {$k$-th object class} ${e}^{\text{o}}_k \in \R^{d_\text{t}}$ after projecting both patch 
and text features into a shared embedding space:
\begin{equation}
    \begin{split}
        s^\text{h}(p) &=\text{cos}\left(\texttt{P}^\text{h}_\text{v}(x(p)),\texttt{P}^\text{h}_\text{t}({e}^\text{h})\right), \\
        s^{\text{o}}_k(p)&=\text{cos}\left(\texttt{P}^\text{o}_\text{v}(x(p)),\texttt{P}^\text{o}_\text{t}({e}^{\text{o}}_k)\right),
    \end{split}
    \label{eq:patch_sim}
\end{equation}
where $\texttt{P}^{\text{h}}_\text{v}$ and $\texttt{P}^\text{o}_\text{v}$ are the patch projection layers, and $\texttt{P}^\text{h}_\text{t}$ and $\texttt{P}^\text{o}_\text{t}$ are the text projection layers.

Then, the patch-level softmax is applied along the patch dimension to produce the patch importance score of the human $\alpha^\text{h}(p) \in \R$ and {$k$-th object class} $\alpha^{\text{o}}_k(p) \in \R$:
\begin{equation}
    \begin{split}
        \alpha^\text{h}(p) &= \frac{\exp(s^\text{h}(p)/\tau_p)}{\sum_{p'} \exp(s^\text{h}(p')/\tau_p)}, \\
        \alpha^{\text{o}}_k(p)&= \frac{\exp(s^{\text{o}}_k(p)/\tau_p)}{\sum_{p'} \exp(s^{\text{o}}_k(p')/\tau_p)},
    \end{split}
    \label{eq:patch_importance}
\end{equation}
where $\tau_p$ refers to the temperature.

Based on these importance weights, we aggregate patch features to obtain spatially grounded representations for human $q^\text{h}\in\R^d$ and {$k$-th object class} $q^\text{o}_k\in\R^d$.
The resulting features $q^\text{h}$ and $q^{\text{o}}_k$ serve as spatial cues, capturing local representations of human and object appearances that facilitate more precise interaction reasoning.
The query construction process can be expressed as:
\begin{equation}
        q^\text{h}=\sum_{p}\alpha^\text{h}(p)x(p), \   q^{\text{o}}_k=\sum_{p}\alpha^{\text{o}}_k(p)x(p), 
    \label{eq:vg_query}
\end{equation}

Finally, we concatenate these two features and feed them into a learnable projection layer $\texttt{P}_{q}:R^{2d}\rightarrow R^d$ to produce a spatially grounded pairwise query $q^\text{ho}_k=\texttt{P}_{q}\big([q^\text{h}; q^{\text{o}}_k]\big)$.
The generated spatially grounded pairwise query captures the spatial cues, which help the model to transfer the reasoning capability from image-level to instance-level.

\noindent\textbf{Interaction Decoder.}
In the interaction decoder, each HO query is used to predict the interaction scores for a human–object pair.
Specifically, the decoder performs cross-attention between spatially grounded pairwise query $q^\text{ho}_k\in \mathbb{R}^{d}$ and the image features $x$, producing the decoded query feature $\bar{q}^\text{ho}_k$.
We then obtain interaction classification scores $s^\text{a}_k\in\R^{N_\text{a}}$ by measuring cosine similarity between $\bar{q}^\text{ho}_k$ and the interaction text embeddings ${e}^\text{a}\in\R^{d_\text{t}}$ extracted by the text encoder $\mathcal{T}$.
Finally, we apply the sigmoid function $\sigma$~\cite{siglip} for multi-label interaction classification, yielding the final interaction scores $\hat{s}^\text{a}_k$.
The process can be formulated as follows:
\begin{equation}
    \begin{split}
        \bar{q}^\text{ho}_k&=\texttt{Att}(q^\text{ho}_k,x,x)\in\R^{ d},\\
        \hat{s}^\text{a}_k &= \sigma(\cos\big(\texttt{P}_\text{a}(\bar{q}^\text{ho}_k), {e}^\text{a}\big)) \in \R^{N_\text{a}}, \\
    \end{split}
    \label{eq:sequential_decoding}
\end{equation}
where $\texttt{P}_\text{a}:\R^{d\rightarrow d_\text{t}}$ is a projection layer that makes the dimensionality of the decoded query feature $\bar{q}^{\text{ho}}_k$ equal to that of the interaction text embedding $e^\text{a}$.

\noindent\textbf{Interactiveness-aware learning.}
Since instance-level localization is unavailable under weak supervision, all human–object class pairs are trained for interaction prediction regardless of whether they are actually present in the image.
This can cause the model to respond to irrelevant regions where the corresponding object is absent in the image, leading to spurious feature learning~\cite{recovermatch} that hinders effective optimization.

To mitigate this issue, we introduce an interactiveness-aware learning scheme that encourages the model to emphasize interactive human and object regions while suppressing irrelevant ones.
Specifically, we first apply a sigmoid function~\cite{siglip} to the patch-level similarity scores to obtain the patch-level interactiveness scores 
for the human, $\hat{s}^\text{h}(p) = \sigma\left(s^\text{h}(p)\right)$, and the $k$-th object class, $\hat{s}^\text{o}_k(p) = \sigma\left(s^\text{o}_k(p)\right)$.
We then take a weighted sum of these interactiveness scores using their corresponding patch importance weights $\alpha^{\{\text{h},\text{o}\}}$ to produce the image-level interactiveness scores for the human ${r}^\text{h}$ and $k$-th object class ${r}^{\text{o}}_k$:
\begin{equation}
    \begin{split}
        {r}^\text{h}=\sum_{p} \alpha^\text{h}(p)\hat{s}^\text{h}(p), \quad
        {r}^{\text{o}}_k=\sum_{p} \alpha^{\text{o}}_k(p)\hat{s}^{\text{o}}_k(p),
    \end{split}
    \label{eq:interactiveness_score}
\end{equation}
Finally, we compute the pairwise interactiveness score for human and $k$-th object as ${r}^\text{ho}_k=({{r}^\text{h}{r}^{\text{o}}_k})^{0.5}\in\R$. 

After obtaining the pairwise interactiveness score ${r}_k^{\text{ho}}$, we integrate it with the interaction score $\hat{s}^\text{a}_k\in\R^{N_\text{a}}$ by broadcasting the interactiveness score over all action dimensions to produce the final HOI classification score $\hat{s}^{\text{hoi}}_k\in\R^{N_\text{a}}$.
Then, we train the model with image-level HOI classes $c^\text{hoi}$ with the focal loss~\cite{focal} as:
\begin{equation}
    \mathcal{L}=\mathcal{L_\text{focal}}(\hat{s}^\text{hoi},c^\text{hoi}), \quad \hat{s}_k^\text{hoi} = \hat{s}_k^\text{a}({{r}_k^\text{ho}})^\gamma,
\label{eq:mul_score}
\end{equation}
where $\gamma$ denotes a scaling factor.

The proposed interactiveness score serves as a gating signal for interaction learning.
Since the interactiveness score is computed from the spatial regions relevant to each human–object pair, the model learns to not only suppress responses to irrelevant regions but also highlight interaction-relevant regions, thereby improving fine-grained interaction reasoning.

\subsection{RegFormer for HOI detection}
\label{sec:4.2}
Our framework can be naturally transferred to HOI detection without any additional training, given only an object detector.
This is achieved through a simple modification, where the construction of \textit{spatially grounded query} in the instance encoder and \textit{interactiveness score} is constrained to specific human and object instances predicted by the object detector.

\noindent\textbf{Instance-aware HO query.}
Unlike ML-Decoder~\cite{ml_decoder,RAM}, our approach constructs queries using instance-level information (\eg, bounding boxes) and performs pairwise decoding directly within the decoder, leading to a substantially more efficient HOI detection.
Specifically, we build query representations with the instance localization information provided by the object detector.
Given a $i$-th human $\tilde{\text{h}}_i=\{\tilde{b}^\text{h}_i,\tilde{s}^\text{h}_i\}$ and $j$-th object $\tilde{\text{o}}_j=\{\tilde{b}^\text{o}_j,\tilde{s}^\text{o}_j,\tilde{c}_j\}$ instance pair from the object detector, we first design a region-aware mask $m\in\R^{h\times w}$ for each human and object instance that filters out pixels outside of the instance region.
For example, in the case of the $i$-th human, the region-aware mask is defined as the indicator function $m^{\tilde{\text{h}}}_i(p)=1$ if $p\in\tilde{b}^\text{h}_i$ and $0$ otherwise.
The object mask ${m}^{\tilde{\text{o}}}_j(p)$ is defined in the same manner, using the corresponding object bounding box $\tilde{b}_j^\text{o}$.
We then apply the logarithm of the region-aware mask to suppress irrelevant regions and obtain the instance-level patch importance scores for each human and object instance, as follows:
\begin{equation}
    \begin{split}
        \textcolor{red}{\alpha^{\tilde{\text{h}}}_i(p)} &= \frac{\exp((s^\text{h}(p)+\log\textcolor{red}{{m}^{\tilde{\text{h}}}_i(p)})/\tau_p)}{\sum_{p'} \exp((s^\text{h}(p')+ \log\textcolor{red}{{m}^{\tilde{\text{h}}}_i(p')})/\tau_p)}, \\
        \textcolor{blue}{\alpha^{\tilde{\text{o}}}_j(p)}&= \frac{\exp(({s}^\text{o}_{\tilde{c}_j}(p)+{\log }\textcolor{blue}{{m}^{\tilde{\text{o}}}_j(p)})/\tau_p)}{\sum_{p'} \exp(({s}^\text{o}_{\tilde{c}_j}(p')+{\log }\textcolor{blue}{{m}^{\tilde{\text{o}}}_j(p')})/\tau_p)}.
    \end{split}
    \label{eq:dense_weighting}
\end{equation}
The instance-aware HO query $q_{ij}^{\widetilde{\text{ho}}}$ is computed by replacing the image-level patch importance ($\alpha^\text{h},\alpha^\text{o}_k$) in \cref{eq:vg_query} with its instance-specific counterpart for each human and object instance. 
The subsequent process in the RegFormer that processes the query ${q}^{\widetilde{\text{ho}}}_{ij}$ remains identical to \cref{eq:sequential_decoding}, resulting in the interaction classification score between $i$-th human and $j$-th object, $\tilde{s}^{{\text{a}}}_{ij}\in\R^{N_\text{a}}$.

\begin{figure}[t!]
    \centering 
    \includegraphics[width=0.95\linewidth]{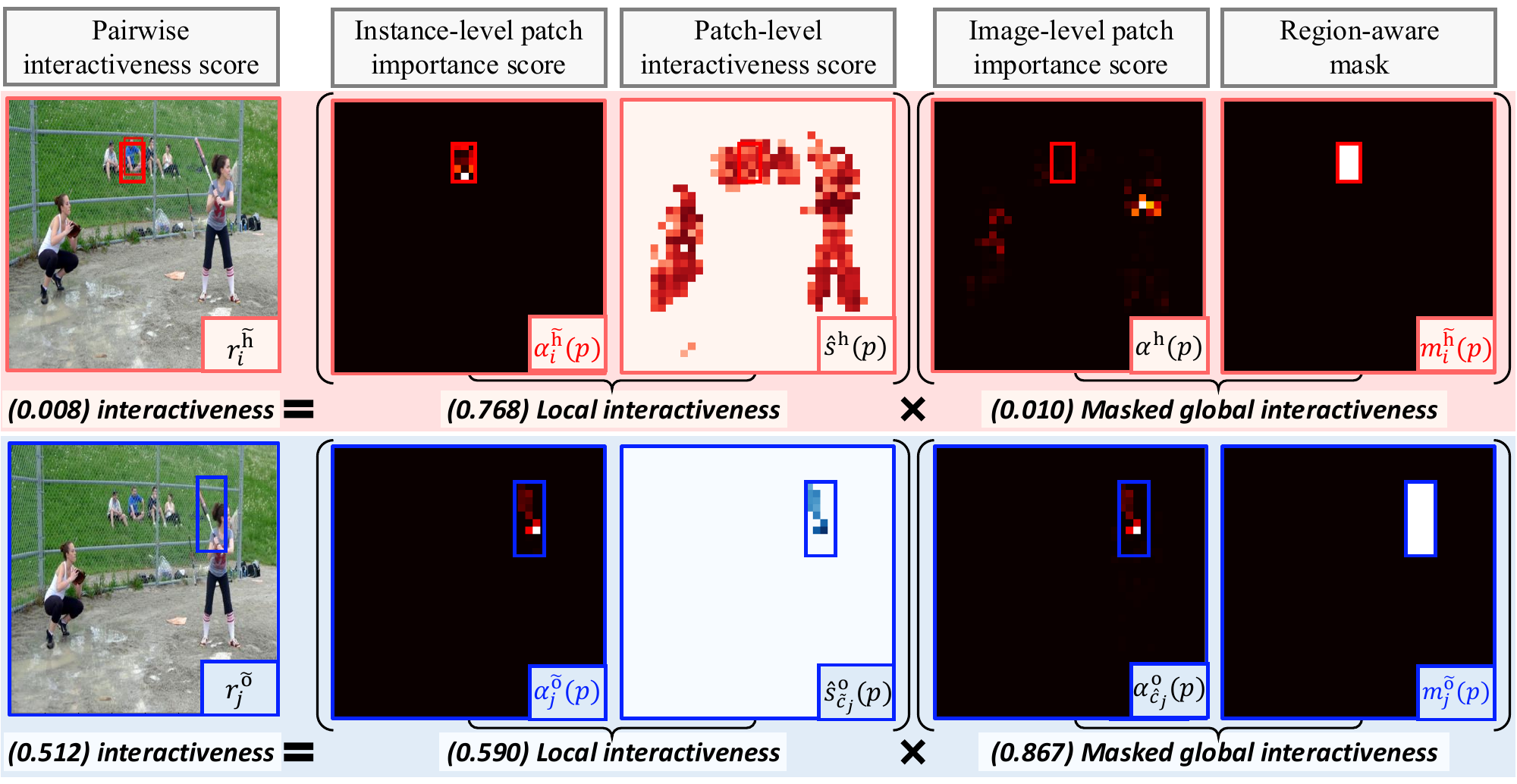}
    \caption{\textbf{Visualization of interactiveness score in HOI detection.}
    {Top row shows the interactiveness score for the human (\textcolor{red}{red box}), and the bottom row shows the score for the object (\textcolor{blue}{blue box}). 
    In row 1, for the human, the masked global interactiveness (0.01) corrects the inflated local interactiveness (0.768) caused by strong semantic alignment between the human and the patch, reducing the pairwise score of the non-interactive region (\textcolor{red}{red box}) to 0.008.
    }
    }
    \label{fig:interactiveness_socring}
\end{figure}
\noindent\textbf{Instance-aware Interactiveness.}
Similar to instance-aware query construction, the image-level interactiveness score (\cref{eq:interactiveness_score}) is extended to an instance-level formulation to assess the interactiveness within provided instance regions.
We first aggregate the patch-level interactiveness scores $\hat{s}(p)$ within the region specified by the region-aware mask, producing a \textit{local interactiveness} score that reflects how strongly an instance exhibits interactive behavior within its own spatial boundary.

However, as shown in the third column of \cref{fig:interactiveness_socring}, this local signal can sometimes yield unexpectedly high scores for non-interactive instances due to strong semantic alignment, even when those regions are identified as the truly salient interactive areas in the image.
To mitigate this, we incorporate the \textit{masked global interactiveness}, derived from the image-level patch importance score $\alpha^{\{\text{h},\text{o}\}}$ within the corresponding instance region.
Unlike the local signal, this term reflects interaction saliency in a global context, which amplifies the contrast between interactive and non-interactive regions (fourth column of \cref{fig:interactiveness_socring}) and thus facilitates the suppression of non-interactive pairs.
By combining the local and global interactiveness priors, the model jointly focuses on both spatially confined and scene-level interactive cues, resulting in a refined interactiveness score for each human ${r}^{\tilde{\text{h}}}$ and object ${r}^{\tilde{\text{o}}}$, which can be formulated as follows:
\begin{equation}
    \begin{split}
        {r}^{\tilde{\text{h}}}_i&=(\sum_{p} \textcolor{red}{\alpha^{\tilde{\text{h}}}_i(p)}\hat{s}^\text{h}(p))(\sum_{p}\alpha^{{\text{h}}}(p)\textcolor{red}{m^{\tilde{\text{h}}}_i(p)}) \\
        {r}^{\tilde{\text{o}}}_j&=\underbrace{(\sum_{p} \textcolor{blue}{\alpha^{\tilde{\text{o}}}_j(p)}\hat{s}^\text{o}_{\tilde{c}_j}(p))}_{\text{local interactiveness}}\underbrace{(\sum_{p}\alpha^{\text{o}}_{\tilde{c}_j}(p)\textcolor{blue}{m^{\tilde{\text{o}}}_j(p)})}_{\text{masked global interactiveness}} \\
    \end{split}
    \label{eq:dense_interactiveness_score}
\end{equation}

\noindent\textbf{Inference.}
Based on the instance-aware query and interactiveness score, the final HOI prediction $\tilde{s}^{{\text{hoi}}}_{ij}$ for each human–object instance pair $(\tilde{\text{h}}_i, \tilde{\text{o}}_j)$ is obtained by combining multiple complementary cues.
Specifically, the instance-aware query $q_{ij}^{\widetilde{\text{ho}}}$ is processed through RegFormer to produce the interaction prediction $\tilde{s}_{ij}^{{\text{a}}}$, which is further modulated by the pairwise interactiveness scores ${r}^{\widetilde{\text{ho}}}_{ij}=({r}^{\tilde{\text{h}}}_i{r}^{\tilde{\text{o}}}_j)^{0.5}$ as well as their detector confidence scores ($\tilde{s}^\text{h}_i$ and $\tilde{s}^\text{o}_j$).
The final prediction is formulated as:
\begin{equation}
\tilde{s}^{{\text{hoi}}}_{ij} = \tilde{s}_{ij}^{{\text{a} }}\cdot ({r}^{\widetilde{\text{ho}}}_{ij})^{\gamma} \cdot (\tilde{s}^\text{h}_i\tilde{s}^\text{o}_j)^{\lambda},
\label{eq:dense_inference}
\end{equation}
where $\lambda$ denotes the scaling factor for the detection score.
\section{Experiments}
\subsection{Experimental setup}
\noindent\textbf{Datasets.}
To evaluate the effectiveness of RegFormer for HOI reasoning, we conduct experiments on two widely used benchmarks: V-COCO~\cite{vcoco} and HICO-DET~\cite{hicodet}.
V-COCO is a subset of MS-COCO~\cite{mscoco}, containing 80 object categories and 29 interaction categories, including 4 body-part actions that do not require an associated object. 
HICO-DET consists of 80 object categories and a more diverse set of 117 interaction categories, yielding 600 valid HOI triplets. 
Notably, HICO-DET exhibits a highly long-tailed class distribution when measured across HOI triplets, making it a challenging benchmark for robust interaction reasoning.
Details of the evaluation metrics for each dataset and setting are provided in the supplementary material.

\noindent\textbf{Implementation details.}
For the vision encoder, we use CLIP~\cite{CLIP} with a ResNet50 backbone (CLIP-RN50) and DINOv2~\cite{oquabdinov2} with ViT-S and ViT-B backbones, referred to as DINO-S and DINO-B, respectively.
For the text encoder, we employ CLIP~\cite{CLIP} using its ResNet50 (CLIP-RN50) and ViT-B (CLIP-B) variants.
During training, all vision and text encoders are frozen to preserve pretrained representations.
We demonstrate the effectiveness of our framework across a broad spectrum of detectors from RCNN-based models~\cite{faster_rcnn,faster_rcnn_V2} to Transformer-based architectures~\cite{detr,hybrid}.
In our experiments and analysis, we adopt DINO-B for vision, CLIP-B for text, and DETR for object detector by default, unless otherwise specified.
Details of the training and inference hyperparameters are provided in the supplementary material.

\subsection{Main results}
\noindent\textbf{Effectiveness of RegFormer.}
\cref{tab:component} presents the component-wise effectiveness of {RegFormer}, evaluated on HOI classification (HICO) and HOI detection (HICO-DET). 
{We analyze three components. {\textbf{HO}$\rightarrow$\textbf{I}} denotes the process of predicting interactions given human and object pairs in the interaction decoder, and \textbf{SG} indicates the Spatially Grounded pairwise query generated by the pairwise instance encoder. \textbf{IA} refers to the InterActiveness scoring. }
The baseline (a) corresponds to ML-Decoder~\cite{ml_decoder,RAM}, which serves as our base architecture.
We use DINO-S as the vision backbone for our component analysis.

In HOI classification (HICO), replacing the original ML-Decoder with a sequential decoding process (HO $\rightarrow$ I) brings a modest gain of {+1.1 mAP}.
Introducing a spatially grounded query further improves performance (+1.8 mAP) by enhancing the semantic alignment between queries and image features, enabling effective relational reasoning.
Incorporating {interactiveness score (IA)} yields an even larger improvement (+3.6 mAP) by preventing feature learning from irrelevant objects, resulting in stable training.
When our query construction and interactiveness-aware learning are combined, the model achieves the highest classification performance of 57.6 mAP, which is an improvement of +5.0 mAP compared to the baseline ML-Decoder, demonstrating the complementary benefits of spatially grounded query construction and interactiveness-aware learning.

When our module is applied to HOI detection (HICO-DET), 
similar to HOI classification, all components consistently improve performance. 
Interestingly, the {SG query} enables a significant performance gain of 4.59 in Full and 5.72 in Rare, compared to the base model (ML-Decoder). 
This is because it still relies on cropped union regions in the image, which limits efficiency and provides no mechanism to filter out irrelevant instances.
Adding {interactiveness scoring} further boosts performance by effectively suppressing non-interactive pairs, thereby improving its precision even further.
Since both components explicitly ground the reasoning on the corresponding instance pairs, only a single backbone forward is required, maintaining high efficiency.
Finally, combining both maximizes their complementarity, achieving an impressive improvement of \textbf{+12.52} on Full over ML-Decoder. 
This clearly demonstrates that RegFormer not only excels at image-level HOI reasoning (classification) but also transfers remarkably well to instance-level HOI reasoning (detection) \emph{without} any additional training.

\begin{table}[!t]
\centering
\caption{\textbf{Contribution of each component in RegFormer.}
Each component is defined as follows: \textbf{HO}$\rightarrow$ \textbf{I} for our reasoning process in the interaction decoder, \textbf{SG} for Spatially Grounded pairwise query, and \textbf{IA} for InterActiveness scoring.
Without these components, the model is equivalent to ML-Decoder~\cite{ml_decoder,RAM}.
}
\label{tab:component}
\vspace{-10pt}
\resizebox{\columnwidth}{!}{
\begin{tabular}{c |
>{\centering\arraybackslash}p{1.0cm} |
>{\centering\arraybackslash}p{1.0cm} |
>{\centering\arraybackslash}p{1.0cm} |
c|ccc|
>{\centering\arraybackslash}p{1.5cm}}
\toprule
\multirow{2}{*}{} &
\multirow{2}{*}{\textbf{HO}$\rightarrow$ \textbf{I}} &
\multirow{2}{*}{\textbf{SG}} &
\multirow{2}{*}{\textbf{IA}} &
\multicolumn{1}{c|}{\cellcolor{hico}\textbf{HICO}} &
\multicolumn{3}{c|}{\cellcolor{hicodet}\textbf{HICO-DET}} &
\multirow{2}{*}{\makecell{\# Backbone\\Forward}} \\
 & & & &
\textbf{mAP} & \textbf{\apf} & \textbf{\apr} & \textbf{\apnr} & \\
\midrule
(a) & \multicolumn{3}{c|}{\cellcolor{gray!30}\textbf{ML-Decoder}~\cite{ml_decoder,RAM}} & 52.6 & 17.49 & 18.19 & 17.28 & $N_{\tilde{\text{h}}}N_{\tilde{\text{o}}}$ \\
(b) & \checkmark &  &  & 53.7 & 17.63 & 18.20 & 17.46 & $N_{\tilde{\text{h}}}N_{\tilde{\text{o}}}$ \\
(c) & \checkmark & \checkmark &  & 54.4 & 22.08 & 23.91 & 21.53 & 1 \\
(d) & \checkmark &  & \checkmark & 56.2 & 23.38 & 24.45 & 23.07 & 1 \\
(e) & \checkmark & \checkmark & \checkmark & \textbf{57.6} & \textbf{30.01} & \textbf{32.05} & \textbf{29.39} & 1 \\
\bottomrule
\end{tabular}
}
\vspace{-20pt}
\end{table}

\noindent\textbf{Comprehensive comparison of HOI detectors.}
Table~\ref{tab:sota} shows the comparison between RegFormer and existing fully supervised and weakly supervised HOI detectors on HICO-DET.
In the weakly supervised setting, RegFormer demonstrates consistent improvements across all detector architectures.
Specifically, when using Faster R-CNN as the detector, RegFormer achieves a +2.19 mAP improvement on Full over the previous state-of-the-art method (Weakly HOI-CLIP) while using the same CLIP-RN50 image–text backbone. 
The gain becomes even larger when employing stronger vision encoders such as DINO-S or DINO-B, which provide richer spatial cues, leading to substantial additional improvements. 

A similar trend is observed when DETR is used as the detector, where performance continually improves as the image and text backbones become more advanced. 
Notably, with DINO-B as the vision backbone and CLIP-B as the text encoder, RegFormer delivers accuracy comparable to fully supervised models and achieves especially strong performance on rare classes, a setting where supervised approaches typically struggle. 
Furthermore, pairing RegFormer with an advanced detector, \eg, $\mathcal{H}$-DETR~\cite{hybrid} leads to a substantial performance improvement, demonstrating strong scalability. 
Note that RegFormer is trained in a detector-agnostic manner, allowing it to be plugged into any detector with a single training process and without detector-specific design~\cite{pvic}.
By avoiding detector proposals during training, it reduces detector biases and prevents error propagation from imperfect detections, benefiting rare classes and further highlighting its practicality.
\begin{table}[t]
\centering
\caption{\textbf{Weakly \& Fully supervised HOI detection on HICO-DET benchmark dataset}. 
}
\label{tab:sota}
\vspace{-10pt}
\small
\setlength{\tabcolsep}{2pt}
\resizebox{\columnwidth}{!}{
\begin{tabular}{l c cc ccc}
\toprule
\multirow{2}{*}{\textbf{Method} }
& \multirow{2}{*}{\textbf{Detector} }
& \multicolumn{2}{c}{\textbf{Backbone}}
& \multirow{2}{*}{\apf} & \multirow{2}{*}{\apr} & \multirow{2}{*}{\apnr} \\
\cmidrule(lr){3-4}
&  & \textbf{Vision} & \textbf{Text} 
&  &  &  \\
\midrule
\rowcolor{gray!10}
\multicolumn{7}{l}{\textbf{\textit{Fully Supervised}}} \\
HOTR~\cite{hotr} & DETR& RN50  & - & 25.10&17.34 &27.42  \\
QPIC~\cite{qpic} & DETR & RN50  & -  & 29.07 & 21.85 & 31.23 \\
GEN-VKLT~\cite{genvlkt} & DETR & RN50  & CLIP-B &33.75 &29.25 &35.10 \\ 
ADA-CM~\cite{ada_cm} & DETR & CLIP-B  & CLIP-B &33.80 &31.72 & 34.42 \\
CMMP~\cite{cmmp} & DETR & CLIP-B  & CLIP-B &32.26& 33.53&33.24  \\
HOICLIP~\cite{hoiclip} & DETR & CLIP-B  & CLIP-B &34.69& 31.12&35.74 \\

\midrule
\rowcolor{gray!10}
\multicolumn{7}{l}{\textbf{\textit{Weakly Supervised}}} \\
Explanation-HOI~\cite{explainhoi} 
& Faster R-CNN & RNeXt101 & - 
& 10.63 & 8.71 & 11.20 \\
MX-HOI~\cite{mxhoi} 
& Faster R-CNN & RN101 & -
& 16.14 & 12.06 & 17.50 \\
OpenCat~\cite{opencat} 
& Faster R-CNN & RN101 & RoBERTa
& 19.72 & 14.56 & 21.01 \\
PPR-FCN~\cite{pprfcn} 
& Faster R-CNN & CLIP-RN50 & -
& 17.55 & 15.69 & 18.41 \\
Weakly HOI-CLIP~\cite{weakhoiclip} 
& Faster R-CNN & CLIP-RN50 & CLIP-RN50
& 22.89 & 22.41 & 23.03 \\
\rowcolor{cyan!10}
RegFormer 
& Faster R-CNN & CLIP-RN50 & CLIP-RN50
& {25.08} & {25.76} & {24.88} \\
\rowcolor{cyan!10}
RegFormer 
& Faster R-CNN & DINO\text{-}S & CLIP-B
& {30.06} & {30.67} & {29.88} \\
\rowcolor{cyan!10}
RegFormer 
& Faster R-CNN & DINO\text{-}B & CLIP-B
& \textbf{33.33} & \textbf{35.04} & \textbf{32.82} \\
\midrule
AlignFormer~\cite{alignformer} 
& DETR & RN50 & -
& 19.26 & 14.00 & 20.83 \\
\rowcolor{cyan!10}
RegFormer 
& DETR & CLIP-RN50 & CLIP-RN50
& 24.89 & 26.86 & 24.31 \\
\rowcolor{cyan!10}
RegFormer 
& DETR & DINO\text{-}S & CLIP-B
& 30.01 & 32.05 & 29.39 \\
\rowcolor{cyan!10}
RegFormer 
& DETR & DINO\text{-}B & CLIP-B
& \textbf{32.90} & \textbf{35.18} & \textbf{32.21} \\
\rowcolor{cyan!10}
RegFormer 
& $\mathcal{H}$-DETR & DINO\text{-}B & CLIP-B
& \textbf{38.14} & \textbf{40.31} & \textbf{37.49} \\
\bottomrule
\end{tabular}}
\end{table}

\begin{table}[t]
\centering
\caption{\textbf{Weakly supervised HOI detection on V-COCO}. $\dagger$ indicates additional large-scale weakly supervised pre-training. }
\label{tab:vcoco}
\vspace{-10pt}
\small
\setlength{\tabcolsep}{4pt}
\resizebox{0.8\columnwidth}{!}{
\begin{tabular}{l c c}
\toprule
\textbf{Method} 
& \textbf{Detector} 
& \textbf{\sctwo} \\
\midrule
AlignFormer~\cite{alignformer}  
& DETR & 14.2 \\
OpenCat~\cite{opencat}
& Faster R-CNN & 15.2 \\
OpenCat$^\dagger$~\cite{opencat}
& Faster R-CNN & 36.1 \\
Weakly HOI-CLIP~\cite{weakhoiclip}
& Faster R-CNN & 48.1 \\
\rowcolor{cyan!10}
RegFormer 
& Faster R-CNN & 56.5 \\
\rowcolor{cyan!10}
RegFormer 
& DETR & \textbf{57.5} \\
\bottomrule
\end{tabular}}
\vspace{-10pt}
\end{table}

Additionally, \cref{tab:vcoco} presents weakly supervised HOI detection results on V-COCO.
We use DINO-S and CLIP-B for image and text encoder in RegFormer, respectively.
RegFormer consistently outperforms prior weakly supervised approaches across both Faster R-CNN and DETR detectors, achieving 56.5 \sctwo \ and 57.5 \sctwo, respectively, and setting a new state-of-the-art, demonstrating its high applicability.

\noindent\textbf{Comparison in zero-shot HOI detection.}
\cref{tab:zero_shot} compares the zero-shot HOI detection performance under two evaluation settings~\cite{fcl}: RF-UC (Rare First Unseen Compositions) and NF-UC (Non-Rare First Unseen Compositions).
Compared to the weakly supervised baseline OpenCat~\cite{opencat}, RegFormer shows consistent and substantial improvements across both settings.
In the NF-UC setting, RegFormer achieves 25.44 / 30.86 / 29.78 for Unseen / Seen / Full, consistently outperforming OpenCat across all splits.
In the more challenging RF-UC setting, where unseen compositions are generally difficult and even fully supervised methods show a sharp drop in unseen performance, RegFormer surpasses OpenCat by a large margin of 10.07 mAP on unseen compositions while maintaining balanced performance across seen and unseen cases, resulting in a significantly higher harmonic mean.
It is worth noting that OpenCat is pretrained on a large-scale dataset of 750k images, yet RegFormer still exhibits stronger generalization to unseen compositions.
These results highlight that RegFormer not only excels in handling unseen object–action compositions but also maintains a strong balance between seen and unseen predictions, validating its robust zero-shot reasoning capability under weak supervision.

\begin{table}[t]
\centering
\caption{\textbf{Zero-shot HOI detection.} 
RF-UC and NF-UC denote rare-first and non-rare-first unseen compositions, respectively.
$^\dagger$ indicates the use of large-scale pretraining.
}
\label{tab:zero_shot}
\vspace{-8pt}
\resizebox{\columnwidth}{!}{
\begin{tabular}{lccccccccc}
\toprule
& \multicolumn{4}{c}{\textbf{RF-UC}} 
& \multicolumn{4}{c}{\textbf{NF-UC}} \\
\cmidrule(lr){2-5} \cmidrule(lr){6-9}
\textbf{Method} 
& $\text{Unseen}$ & $\text{Seen}$ & $\text{Full}$ & $\text{HM}$ 
& $\text{Unseen}$ & $\text{Seen}$ & $\text{Full}$ & $\text{HM}$ \\
\midrule
\rowcolor{gray!10}
\multicolumn{9}{l}{\textbf{\textit{Fully Supervised}}} \\
FCL~\cite{fcl} 
& 13.16 & 24.23 & 22.01 & 17.06
& 18.66 & 19.55 & 19.37 & 19.09 \\
ATL~\cite{atl} 
& 9.18 & 24.67 & 21.57 & 13.38
& 18.25 & 18.78 & 18.67 & 18.51 \\
RLIP~\cite{rlip} 
& 19.19 & 33.35 & 30.52 & 24.36
& 20.27 & 27.67 & 26.19 & 23.40 \\
GEN-VLKT~\cite{genvlkt} 
& 21.36 & 32.91 & 30.56 & 25.91
& 25.05 & 23.38 & 23.71 & 24.19 \\
EoID~\cite{eoid} 
& 22.04 & 31.39 & 29.52 & 25.90
& 26.77 & 26.66 & 26.69 & 26.71 \\
ADA-CM~\cite{ada_cm} 
& 27.63 & 34.35 & 33.01 & 30.63
& 32.41 & 31.13 & 31.39 & 31.76 \\
HOICLIP~\cite{hoiclip} 
& 25.53 & 34.35 & 32.99 & 29.29
& 26.39 & 28.10 & 27.75 & 27.22 \\
CLIP4HOI~\cite{clip4hoi} 
& 28.47 & 35.48 & 34.06 & 31.59
& 31.44 & 28.26 & 28.90 & 29.77 \\
CMMP~\cite{cmmp} 
& 29.45 & 32.87 & 32.18 & 31.07
& 32.09 & 29.71 & 30.18 & 30.85 \\
\midrule
\rowcolor{gray!10}
\multicolumn{9}{l}{\textbf{\textit{Weakly \& Fully Supervised}}} \\
OpenCat$^\dagger$~\cite{opencat}    
& 21.46 & \textbf{33.86} & \textbf{31.38} & 26.27
& 23.25 & 28.04 & 27.08 & 25.42 \\
\midrule
\rowcolor{gray!10}
\multicolumn{9}{l}{\textbf{\textit{Weakly Supervised}}} \\
\rowcolor{cyan!10}
RegFormer 
& \textbf{31.53} & 30.25 & 30.50 & \textbf{30.88}
& \textbf{25.44} & \textbf{30.86} & \textbf{29.78} & \textbf{27.89} \\
\bottomrule
\end{tabular}}
\vspace{-6mm}
\end{table}

\begin{table}[ht]
\centering
\caption{\textbf{Ablation on Interactiveness Scoring on HICO-DET}.
\textit{Local} and \textit{Global} stand for local interactiveness and masked global interactiveness, respectively.
}
\label{tab:interactiveness_ablation}
\vspace{-8pt}
\small
\begin{tabular}{ccccc}
\toprule
\textbf{Local} & \textbf{Global} & \apf & \apr & \apnr \\
\midrule
\rowcolor{gray!10}
\multicolumn{5}{l}{\textbf{\textit{Without Interactiveness-aware learning}}} \\
& & 22.08 & 23.91 & 21.53 \\
 & \checkmark & 26.02 & 26.81 & 25.79 \\
 \rowcolor{gray!10}
\multicolumn{5}{l}{\textbf{\textit{With Interactiveness-aware learning}}} \\
\checkmark & & 23.44 & 25.77 & 22.75 \\
\checkmark & \checkmark & \textbf{30.01}& \textbf{32.05} & \textbf{29.39} \\
\bottomrule
\end{tabular}
\vspace{-5mm}
\end{table}

\subsection{Ablation and analysis.}

\noindent\textbf{Effect of Interactiveness Scoring components.}
\cref{tab:interactiveness_ablation} shows the effect of the proposed local and masked global interactiveness scoring on HICO-DET.
We use DINO-S and CLIP-B for vision and text encoder, respectively.
The results show that the two components are complementary.
Local interactiveness provides pair-specific localized cues, while masked global interactiveness is more effective at suppressing non-interactive instances.
Their combination with interactiveness-aware learning yields the largest gain, indicating that local and global interactiveness jointly improve pairwise interaction reasoning.
\cref{fig:interactiveness_socring} provides a qualitative illustration of each component.

\noindent\textbf{Qualitative analysis.}
\begin{figure}[t]
    \centering
    \includegraphics[width=0.9\linewidth]{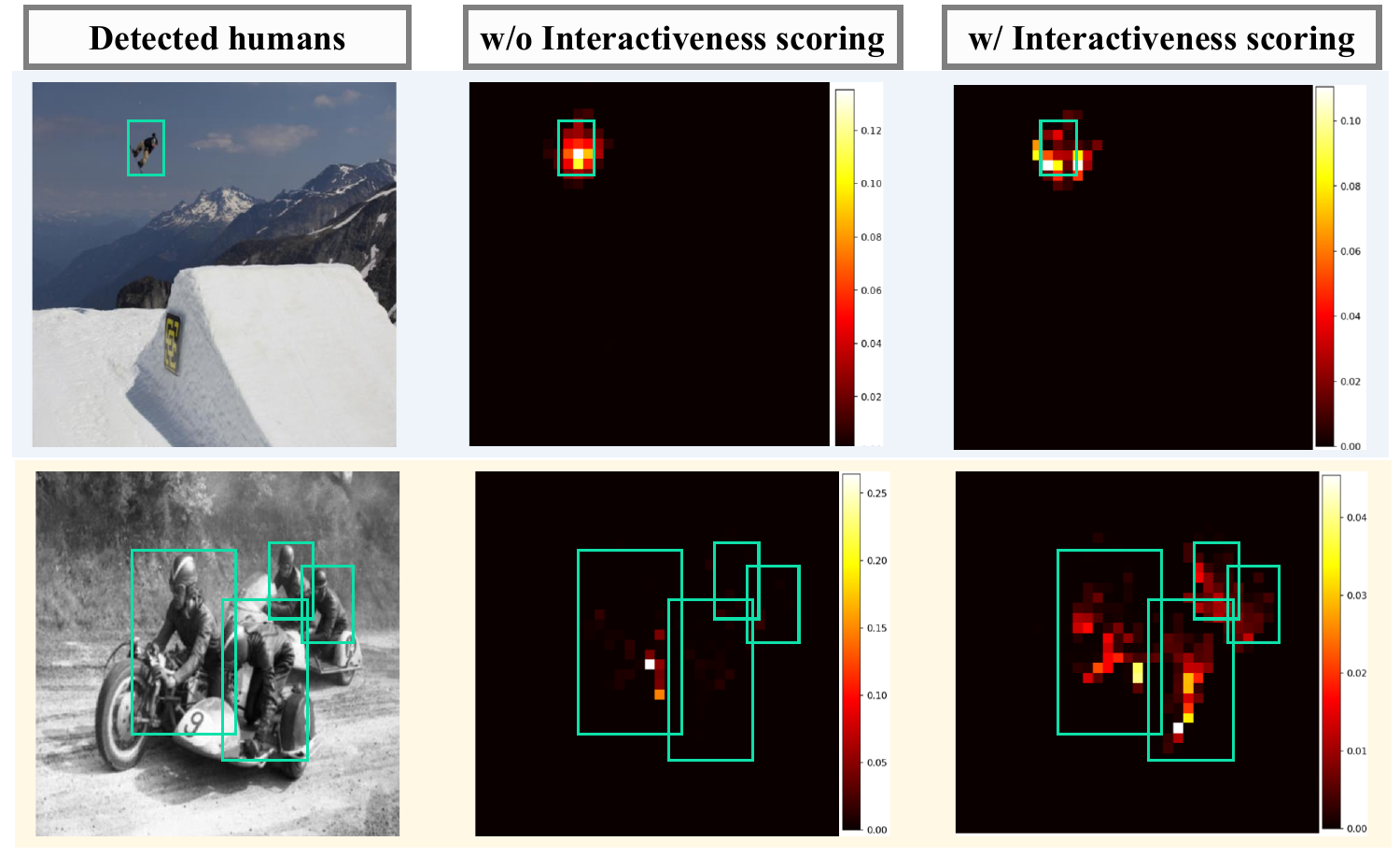}
    \vspace{-0.2cm}
    \caption{\textbf{Qualitative results on patch importance score}.
    We visualize the human patch importance score, $\alpha^\text{h}(p)$.
    }
    \label{fig:qual_patch}
    \vspace{-8pt}
\end{figure}
We analyze how interactiveness-aware learning affects the patch importance scores, focusing on the human patch importance scores $\alpha^\text{h}(p)$ as shown in \cref{fig:qual_patch}.
Both settings use spatially grounded queries.
In sparse scenes (row 1), where a single human appears, the model highlights human regions even without interactiveness-aware learning, indicating that spatial grounding alone can capture local cues.
However, in dense scenes (row 2), this implicit learning fails to localize all interactive individuals.
With interactiveness-aware learning, explicit signals about human existence lead to more accurate and consistent localization across multiple interactive humans.
Overall, these results demonstrate that interactiveness-aware learning effectively elicits the interaction between instances without instance-level annotations.

\section{Conclusion}
We propose \textbf{RegFormer}, a novel interaction reasoning module that unifies image-level and instance-level HOI reasoning within a single framework.
RegFormer adopts a sequential decoder architecture that strengthens locality-aware learning through spatially grounded query construction and interactiveness-aware supervision.
Owing to this design, the model can be directly transferred to HOI detection at inference time by incorporating a region-aware adjustment, without any additional training, while enabling highly efficient pairwise reasoning with only a single backbone forward.
Experimental results show that, although trained with only image-level supervision, RegFormer exceeds the performance of prior weakly supervised methods and approaches fully supervised counterparts in some settings, highlighting its strong generalization and scalability.

\section*{Acknowledgments}
This work was partly supported by Korea Research Institute for defense Technology planning and advancement - Grant funded by Defense Acquisition Program Administration (DAPA)(KRIT-CT-23-021) (50\%), the National Research Foundation of Korea (NRF) grant funded by the Korea government (MSIT) (NRF-2023R1A2C2005373) (25\%), and the InnoCORE program of the Ministry of Science and ICT (N10250156) (25\%).

{
    \small
    \bibliographystyle{unsrt}
    \bibliography{main}
}


\clearpage

\appendix
\section*{Appendix}
This supplementary material provides additional details beyond the main paper, organized as follows: \\

\noindent\textbf{(A) Details of experimental settings}
\begin{itemize}
  \item[] -- (\ref{metrics}) Evaluation metrics
  \item[] -- (\ref{imple}) Implementation details
\end{itemize}

\noindent\textbf{(B) Additional ablations and analyses}
\begin{itemize}
  \item[] -- (\ref{subsec:efficiency_comparison}) Efficiency comparison
  \item[] -- (\ref{subsec:scaling_factor_sen}) Sensitivity analysis on scaling factors
  \item[] -- (\ref{subsec:sensitivity_tau}) Ablation on temperature scaling for patch importance score
  \item[] -- (\ref{subsec:backbone_impact}) Impact of vision backbone
  \item[] -- (\ref{subsec:qual_results}) Qualitative results on false positive suppression
\end{itemize}

  \section{Details of experimental settings.}
\subsection{Evaluation metrics}
\label{metrics}
Our work is evaluated on two HOI benchmarks, HICO-DET~\cite{hicodet} and V-COCO~\cite{vcoco}.
For \textbf{HICO-DET} evaluation, we compute the average precision (AP) for each HOI class and group them by annotation frequency; classes with fewer than 10 samples are defined as \textit{rare} (\textbf{\apr}), and the others as \textit{non-rare} (\textbf{\apnr}).
The mean average precision over all classes is reported as \textbf{\apf}.
We additionally evaluate our model under the zero-shot setting~\cite{fcl} on HICO-DET, where a subset of HOI classes is held out during training. 
Following the protocol in~\cite{fcl}, we evaluate under two settings: (1) \textbf{RF-UC} (rare-first unseen composition), where the rare classes in HICO-DET are held out during training, and (2) \textbf{NF-UC} (non-rare-first unseen composition), where the non-rare classes are withheld during training.
For each setting, we report performance on \textbf{Unseen}, \textbf{Seen}, and \textbf{Full} splits, as well as the harmonic mean (\textbf{HM}) between unseen and seen results to assess their balance.
For \textbf{V-COCO}, following the current evaluation protocol~\cite{upt}, we report the average precision (\textbf{\sctwo}) over the 24 action categories out of the 29 annotated ones, excluding the 4 body-part actions and the `point-obj' class.

\subsection{Implementation details}
\label{imple}

During training RegFormer in the weakly supervised setting, we use a learning rate of $2\times10^{-4}$ with a cosine scheduler for 5 epochs.
The batch size is set to 32 for HICO-DET and 16 for V-COCO.
The embedding dimension $d$ is aligned with the backbone feature dimension $d_\text{v}$, temperature $\tau_p$ is set to 0.05, and $\gamma$ is set to 1.0.
For normalizing the interaction classification score $\hat{s}^\text{a}$, and the interactiveness scores of human $\hat{s}^\text{h}$ and object $\hat{s}^\text{o}$, we use sigmoid with a learnable temperature and bias, which are initialized to $\log 10$ and $-5$, respectively, following SigLIP~\cite{siglip}.
The input resolution follows the vision backbone’s pre-training setting.

To perform HOI detection, our framework first retrieves all possible human and object proposals from an off-the-shelf object detector.
For object detection proposals, we follow~\cite{ada_cm} by filtering out detections with confidence scores below 0.2 and selecting between 3 and 15 human and object instances, while setting the detection probability scaling factor $\lambda$ to 2.0 for V-COCO and 0.5 for HICO-DET.
\paragraph{Hardware setup}
All training experiments are performed using two RTX 3090 GPUs, while inference time is measured on a single RTX 3090 GPU.

\section{Additional ablations and analyses}

\subsection{Efficiency comparison}\label{subsec:efficiency_comparison}
\begin{figure}[ht]
    \centering
    \includegraphics[width=0.9\linewidth]{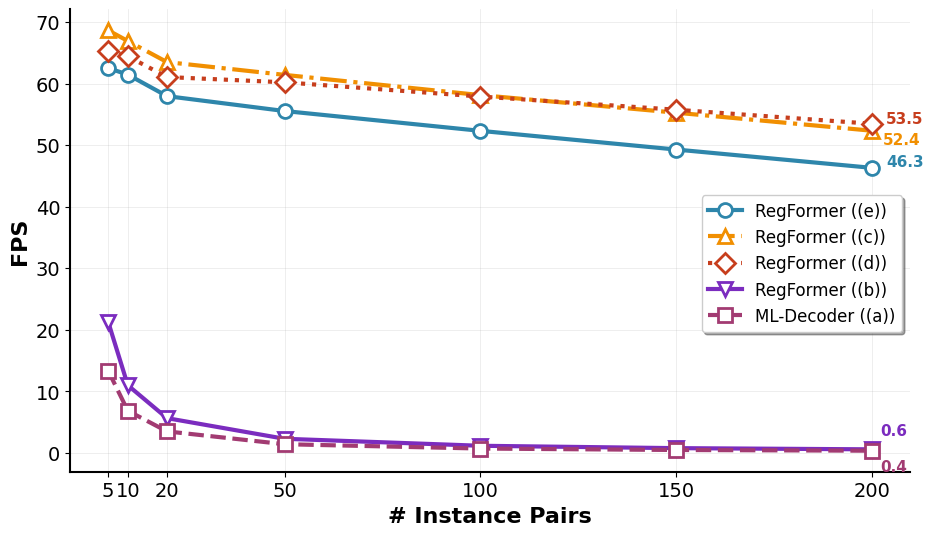}
    \caption{\textbf{Efficiency comparison.}
    (a) ML-Decoder baseline, 
    (b) adding our HO$\rightarrow$I reasoning,
    (c) (b) + spatially grounded pairwise query,
    (d) (b) + interactiveness scoring,
    (e) full RegFormer with all components.}
    \label{fig:fps}
\end{figure}

In \cref{fig:fps}, we compare the efficiency of our model by measuring FPS for each configuration obtained by incrementally adding components on top of the ML-Decoder baseline. 
The evaluated variants correspond to those in Table 1 of the main paper: (a) ML-Decoder baseline, (b) adding HO$\rightarrow$I reasoning, (c) (b) + spatially grounded pairwise query, (d) (b) + interactiveness scoring, and (e) the full RegFormer with all components.
FPS is measured on a single RTX 3090 with batch size 1, averaged over 200 iterations.

Transitioning from (a) to (b) reduces the attention complexity from scaling with the number of HOI triplet classes to the number of object classes, yielding faster inference for images with few instance pairs. 
However, because it still relies on union-cropped regions, its speed gain diminishes as the number of pairs increases. 
Introducing spatially grounded pairwise queries in (c) removes the need for cropped regions and enables pairwise interaction reasoning within a single backbone forward pass, achieving 52.5 FPS even with 200 instance pairs. 
Interactiveness scoring in (d) also supports pairwise processing within a single forward pass and achieves comparable speed.
The full RegFormer in (e) incorporates both components and maintains 46.3 FPS even under a large number of instance pairs, demonstrating consistently high efficiency. 
Overall, these results confirm that RegFormer performs HOI detection with substantially improved computational efficiency compared to ML-Decoder.

\subsection{Sensitivity analysis on scaling factors}
\label{subsec:scaling_factor_sen}
\begin{figure}[ht]
\centering
\includegraphics[width=0.9\linewidth]{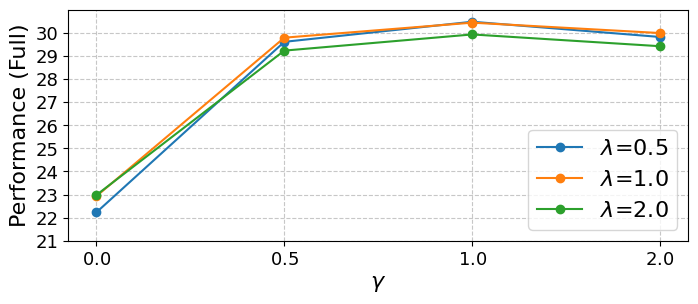}
\caption{Sensitivity analysis on scaling factors.}
\label{fig:score_factor_sensitivity_analysis}
\end{figure}

We conduct a sensitivity analysis on the scaling factor for detector score $\lambda$ and interactiveness score $\gamma$ on HICO-DET, as shown in \cref{fig:score_factor_sensitivity_analysis}.
As long as the scaling factors are set to positive values ($\lambda, \gamma > 0$), the performance remains stable over a wide range of settings, with only minor variations (29.23 to 30.48).

\subsection{Ablation on temperature scaling for patch importance score}\label{subsec:sensitivity_tau}
\begin{table}[ht]
\centering
\caption{\textbf{Sensitivity analysis of $\tau_p$.}}
\label{tab:taup_sensitivity}
\setlength{\tabcolsep}{10pt}
\resizebox{0.85\columnwidth}{!}{
\begin{tabular}{c | c | ccc}
\toprule
\multirow{2}{*}{$\tau_p$} &
\multicolumn{1}{c|}{\cellcolor{hico}\textbf{HICO}} &
\multicolumn{3}{c}{\cellcolor{hicodet}\textbf{HICO-DET}} \\
& \textbf{mAP} & \textbf{\apf} & \textbf{\apr} & \textbf{\apnr} \\
\midrule
0.01&	56.4	&27.32&	29.44&	26.68 \\
0.05&	\textbf{57.6}&	\textbf{30.01}&	\textbf{32.05}&	\textbf{29.39}\\
0.1&	\textbf{57.6}&	29.95&	31.89	&\textbf{29.39}\\
1.0	&53.1&	22.75&	24.78&	22.14\\
\bottomrule
\end{tabular}
}
\end{table}

\cref{tab:taup_sensitivity} reports the sensitivity of our method to the temperature $\tau_p$ used in computing the patch importance score $\alpha^{\{\text{h},\text{o}\}}$, showing stable performance for $\tau_p \in [0.05, 0.1]$ with the best results at $\tau_p = 0.05$.

\subsection{Impact of vision backbone}\label{subsec:backbone_impact}
\begin{table}[ht]
\centering
\caption{\textbf{Performance comparison across different vision backbones.} DETR~\cite{detr} is used to extract instance proposals.}
\label{tab:vision_backbone}
\setlength{\tabcolsep}{10pt}
\begin{tabular}{l|ccc}
\toprule
\textbf{Backbone} $\mathcal{V}$ & \textbf{Full} & \textbf{Rare} & \textbf{Non-rare} \\
\midrule
CLIP-RN50      & 24.89 & 26.86 & 24.31 \\
CLIP-B       & 28.00 & 29.99 & 27.40 \\
SigLIP-B       & 30.84 & 35.62 & 29.41 \\
DINO-B & 32.90 & 35.18 & 32.21 \\

\bottomrule
\end{tabular}
\end{table}

\cref{tab:vision_backbone} presents additional results on HICO-DET using diverse vision backbones $\mathcal{V}$. For CLIP and SigLIP, we use their corresponding text encoders $\mathcal{T}$ that were jointly aligned during multimodal pretraining, while DINO-B is paired with the CLIP-B text encoder. 
SigLIP employs a ViT-B/16 architecture.

\begin{figure*}[ht]
    \centering
    \includegraphics[width=0.9\textwidth]{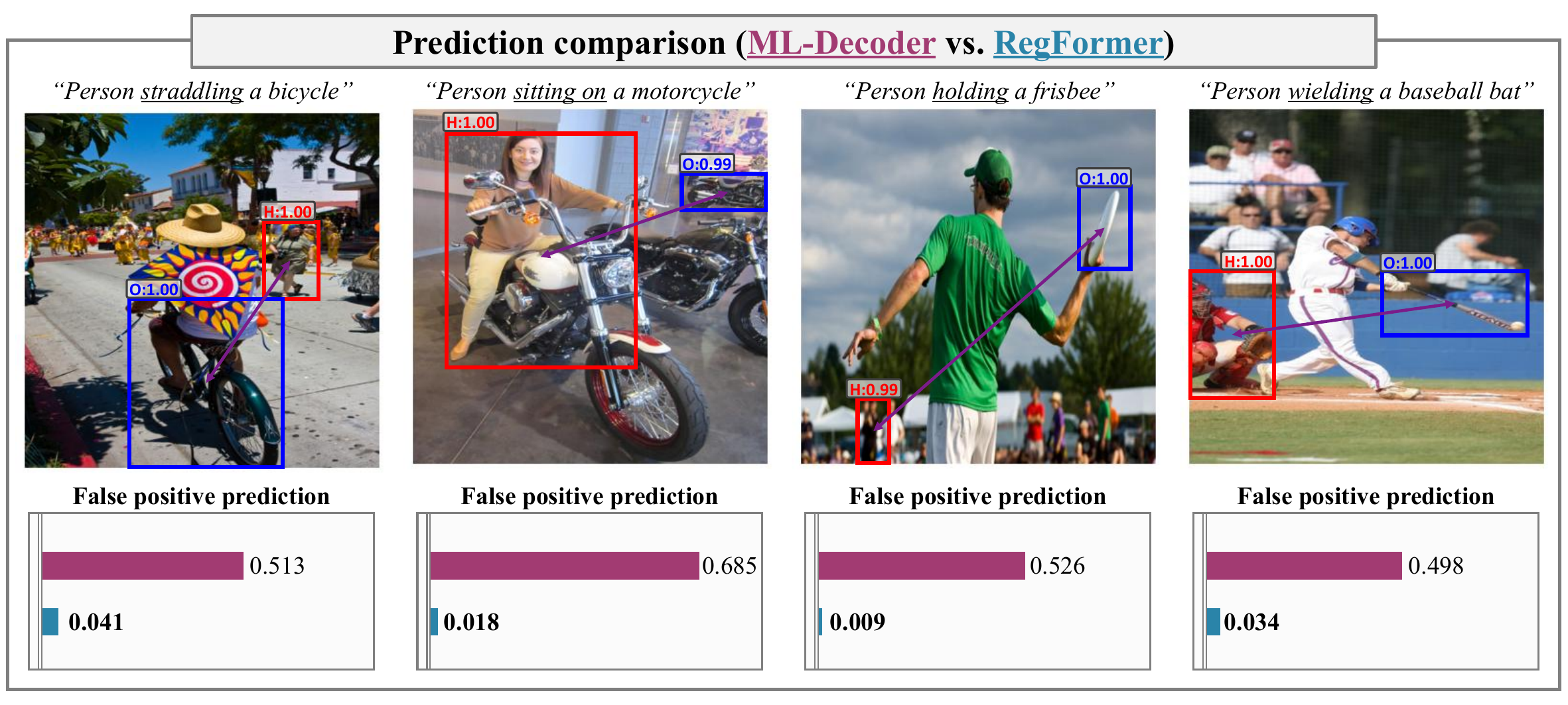}
    \caption{\textbf{Prediction comparison on non-interactive pairs.}}
    \label{fig:suppression_qual}
\end{figure*}

The strong performance achieved with DINO-B suggests that our module benefits greatly from vision backbones with rich local representations~\cite{oquabdinov2}, which help elicit fine-grained object and interaction cues that are crucial for precise local reasoning.

\subsection{Qualitative results on false positive suppression}\label{subsec:qual_results}
\cref{fig:suppression_qual} provides qualitative comparisons of how effectively each model suppresses false positives on non-interactive HOI pairs. 
ML-Decoder, which serves as our baseline, takes the union-cropped region as input image and often assigns high interaction scores to unrelated pairs due to surrounding irrelevant instances. 
In contrast, RegFormer consistently produces near-zero scores for non-interactive pairs, demonstrating its strong ability to suppress false positives and perform more reliable HOI detection.
\newpage

\end{document}